\documentclass[runningheads]{llncs}
\usepackage{graphicx}
\usepackage{color}
\usepackage{tikz}
\usepackage{comment}
\usepackage{amsmath,amssymb,amsfonts}
\usepackage{algorithmic}
\usepackage{graphicx}
\usepackage{textcomp}
\usepackage{xcolor}
\usepackage{epsfig}
\usepackage{graphicx}
\usepackage{algorithm}
\usepackage{algorithmic}
\usepackage{soul}
\usepackage{cool}
\usepackage{float}
\usepackage{authblk}
\usepackage{hyperref}

\begin{document}
\pagestyle{headings}
\mainmatter

\title{Spike-FlowNet: Event-based Optical Flow Estimation with Energy-Efficient Hybrid Neural Networks} 

\titlerunning{Spike-FlowNet: Event-based Optical Flow Estimation}
\authorrunning{C. Lee et al.}


\author{Chankyu Lee\inst{1} \and
Adarsh Kumar Kosta\inst{1} \and
Alex Zihao Zhu\inst{2} \and
Kenneth Chaney\inst{2} \and
Kostas Daniilidis\inst{2} \and
Kaushik Roy\inst{1}
}

\institute{
Purdue University, West Lafayette IN, 47907, USA\\ \email{\{lee2216,akosta,kaushik\}@purdue.edu} \and
University of Pennsylvania, Philadelphia PA, 19104, USA\\ \email{\{alexzhu,chaneyk,kostas\}@seas.upenn.edu}}

\maketitle


\makeatletter{\renewcommand*{\@makefnmark}{}
\footnotetext{The code is publicly available at: \url{https://github.com/chan8972/Spike-FlowNet}}\makeatother}

\makeatletter{\renewcommand*{\@makefnmark}{}
\footnotetext{Associated video: \url{https://youtu.be/8t9xeOLLjL4}}\makeatother}

\begin{abstract}
Event-based cameras display great potential for a variety of tasks such as high-speed motion detection and navigation in low-light environments where conventional frame-based cameras suffer critically. This is attributed to their high temporal resolution, high dynamic range, and low-power consumption. However, conventional computer vision methods as well as deep Analog Neural Networks (ANNs) are not suited to work well with the asynchronous and discrete nature of event camera outputs. Spiking Neural Networks (SNNs) serve as ideal paradigms to handle event camera outputs, but deep SNNs suffer in terms of performance due to the spike vanishing phenomenon. To overcome these issues, we present Spike-FlowNet, a deep hybrid neural network architecture integrating SNNs and ANNs for efficiently estimating optical flow from sparse event camera outputs without sacrificing the performance. The network is end-to-end trained with self-supervised learning on Multi-Vehicle Stereo Event Camera (MVSEC) dataset. Spike-FlowNet outperforms its corresponding ANN-based method in terms of the optical flow prediction capability while providing significant computational efficiency.
\keywords{Event-based Vision \and Optical Flow Estimation \and Hybrid Network \and Spiking Neural Network \and Self-supervised Learning}
\end{abstract}

\section{Introduction}

The dynamics of biological species such as winged insects serve as prime sources of inspiration for researchers in the field of neuroscience, machine learning as well as robotics. The ability of winged insects to perform complex, high-speed maneuvers effortlessly in cluttered environments clearly highlights the efficiency of these resource-constrained biological systems \cite{flymotion2010}. The estimation of motion patterns corresponding to spatio-temporal variations of structured illumination - commonly referred to as optical flow, provides vital information for estimating ego-motion and perceiving the environment. Modern deep Analog Neural Networks (ANNs) aim to achieve this at the cost of being computationally intensive, placing significant overheads on current hardware platforms. A competent methodology to replicate such energy efficient biological systems would greatly benefit edge-devices with computational and memory constraints 
(Note, we will be referring to standard deep learning networks as Analog Neural Networks (ANNs) due to their analog nature of inputs and computations. This would help to distinguish them from Spiking Neural Networks (SNNs), which involve discrete spike-based computations).

Over the past years, the majority of optical flow estimation techniques relied on images from traditional frame-based cameras, where the input data is obtained by sampling intensities on the entire frame at fixed time intervals irrespective of the scene dynamics. Although sufficient for certain computer vision applications, frame-based cameras suffer from issues such as motion blur during high speed motion, inability to capture information in low-light conditions, and over- or under-saturation in high dynamic range environments. 

Event-based cameras, often referred to as bio-inspired silicon retinas, overcome these challenges by detecting log-scale brightness changes asynchronously and independently on each pixel-array element \cite{dvs128}, similar to retinal ganglion cells. Having a high temporal resolution (in the order of microseconds) and a fraction of power consumption compared to frame-based cameras make event cameras suitable for estimating high-speed and low-light visual motion in an energy-efficient manner. However, because of their fundamentally different working principle, conventional computer vision as well as ANN-based methods become no longer effective for event camera outputs. This is mainly because these methods are typically designed for pixel-based images relying on photo-consistency constraints, assuming the color and brightness of object remain the same in all image sequences. Thus, the need for development of handcrafted-algorithms for handling event camera outputs is paramount.

SNNs, inspired by the biological neuron model, have emerged as a promising candidate for this purpose, offering asynchronous computations and exploiting the inherent sparsity of spatio-temporal events (spikes). The Integrate and Fire (IF) neuron is one spiking neuron model \cite{burkitt2006review}, which can be characterized by an internal state, known as the membrane potential. The membrane potential accumulates the inputs over time and emits an output spike whenever it exceeds a set threshold. This mechanism naturally encapsulates the event-based asynchronous processing capability across SNN layers, leading to energy-efficient computing on specialized neuromorphic hardware such as IBM's TrueNorth \cite{merolla2014million} and Intel's Loihi \cite{loihi2018}. However, recent works have shown that the number of spikes drastically vanish at deeper layers, leading to performance degradations in deep SNNs \cite{lee2020enabling}. Thus, there is a need for an efficient hybrid architecture, with SNNs in the initial layers, to exploit their compatability with event camera outputs while having ANNs in the deeper layers in order to retain performance.

In regard to this, we propose a deep hybrid neural network architecture, accommodating SNNs and ANNs in different layers, for energy efficient optical flow estimation using sparse event camera data. To the best of our knowledge, this is the first SNN demonstration to report the state-of-art performance on event-based optical flow estimation, outperforming its corresponding fully-fledged ANN counterpart. 

The main contributions of this work can be summarized as:
\begin{itemize}
    \item We present an input representation that efficiently encodes the sequences of sparse outputs from event cameras over time to preserve the spatio-temporal nature of spike events.
    \item We introduce a deep hybrid architecture for event-based optical flow estimation referred to as Spike-FlowNet, integrating SNNs and ANNs in different layers, to efficiently process the sparse spatio-temporal event inputs. 
    \item We evaluate the optical flow prediction capability and computational efficiency of Spike-FlowNet on the Multi-Vehicle Stereo Event Camera dataset (MVSEC)~\cite{zhu2018multivehicle} and provide comparison results with current state-of-the-art approaches.
\end{itemize}

The following contents are structured as follows. In Section 2, we elucidate the related works. In Section 3, we present the methodology, covering essential backgrounds on the spiking neuron model followed by our proposed input event (spike) representation. This section also discusses the self-supervised loss, Spike-FlowNet architecture, and the approximate backpropagation algorithm used for training. Section 4 covers the experimental results, including training details and evaluation metrics. It also discusses the comparison results with the latest works in terms of performance and computational efficiency.

\section{Related Work}
In recent years, there have been an increasing number of works on estimating optical flow by exploiting the high temporal resolution of event cameras. In general, these approaches have either been adaptations of conventional computer vision methods or modified versions of deep ANNs to encompass discrete outputs from event cameras.

For computer vision based solutions to estimate optical flow, gradient-based approaches using the Lucas-Kanade algorithm~\cite{lucaskanade} have been highlighted in~\cite{benosman1,brosch2015}. Further, plane fitting approaches by computing the slope of the plane for estimating optical flow have been presented in~\cite{benosman2,aung2018}. In addition, bio-inspired frequency-based approaches have been discussed in~\cite{barranco2015}. Finally, correlation-based approaches are presented in~\cite{zhu2017,gallego2018} employing convex optimization over events. In addition,~\cite{liu2018} interestingly uses an adaptive block matching technique to estimate sparse optical flow. 

For deep ANN-based solutions, optical flow estimation from frame-based images has been discussed in Unflow~\cite{meister2018unflow}, which utilizes a U-Net~\cite{unet} architecture and computes a bidirectional census loss in an unsupervised manner with an added smoothness term. This strategy is modified for event camera outputs 
in EV-FlowNet~\cite{zhu2018ev} incorporating a self-supervised loss based on gray images as a replacement for ground truth. Other previous works employ various modifications to the training methodology, such as~\cite{jason2016back}, which imposes certain brightness constancy and smoothness constraints to train a network and~\cite{lai2017} which adds an adversarial loss over the standard photometric loss. In contrast, \cite{zhu2019unsupervised} presents an unsupervised learning approach using only event camera data to estimate optical flow by accounting for and then learning to rectify the motion blur. 

All the above strategies employ ANN architectures to predict the optical flow. However, event cameras produce asynchronous and discrete outputs over time, and SNNs can naturally capture their spatio-temporal dynamics, which are embedded in the precise spike timings. Hence, we posit that SNNs are suitable for handling event camera outputs. Recent SNN-based approaches for event-based optical flow estimation include \cite{Orchard2013,haessig2018spiking,paredes2019unsupervised}. Researchers in \cite{Orchard2013} presented visual motion estimation using SNNs, which accounts for synaptic delays in generating motion-sensitive receptive fields. In addition, \cite{haessig2018spiking} demonstrated real-time model-based optical flow computations on TrueNorth hardware for evaluating patterns including rotating spirals and pipes. Authors of \cite{paredes2019unsupervised} presented a methodology for optical flow estimation using convolutional SNNs based on Spike-Time-Dependent-Plasticity (STDP) learning \cite{diehl2015unsupervised}. The main limitation of these works is that they employ shallow SNN architectures, because deep SNNs suffer in terms of performance. Besides, the presented results are only evaluated on relatively simple tasks. 
In practice, they do not generally scale well to complex and real-world data, such as that presented in MVSEC dataset \cite{zhu2018multivehicle}. In view of these, a hybrid approach becomes an attractive option for constructing deep network architectures, leveraging the benefits of both SNNs and ANNs.

\section{Method} 
\subsection{Spiking Neuron Model}
\begin{figure}[h]
\begin{center}
\includegraphics[width=0.74\columnwidth]{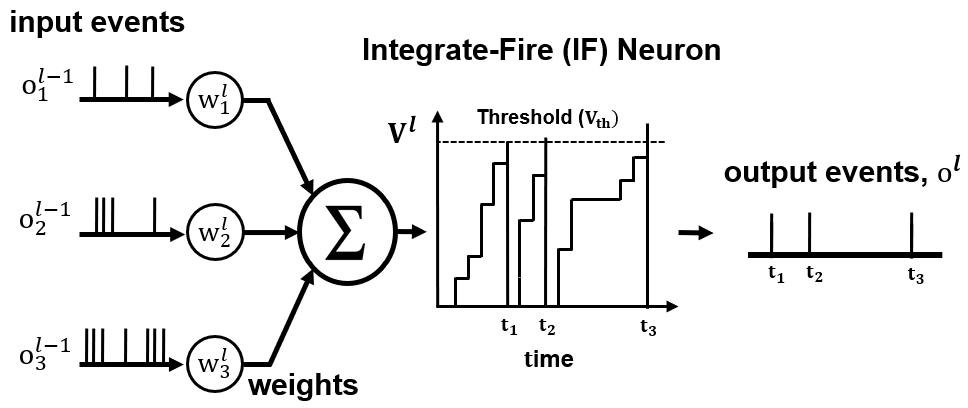}
\caption{The dynamics of an Integrate and Fire (IF) neuron. The input events are modulated by the synaptic weight to be integrated as the current influx in the membrane potential. Whenever the membrane potential crosses the threshold, the neuron fires an output spike and resets the membrane potential.}
\label{fig1}
\end{center}
\vspace{-8mm}
\end{figure}

The spiking neurons, inspired by biological models \cite{dayan2001theoretical}, are computational primitives in SNNs. We employ a simple IF neuron model, which transmits the output signals in the form of spike events over time. The behavior of IF neuron at the $l^{th}$ layer is illustrated in Fig. \ref{fig1}. The input spikes are weighted to produce an influx current that integrates into neuronal membrane potential ($V^l$).
\begin{equation}
V^l[n+1] = V^l[n] + w^{l}*o^{l-1}[n]
\label{eq1}
\end{equation}
where $V^l[n]$ represents the membrane potential at discrete time-step $n$, $w^l$ represents the synaptic weights and $o^{l-1}[n]$ represents the spike events from the previous layer at discrete time-step $n$. When the membrane potential overcomes the firing threshold, the neuron emits an output spike and resets the membrane potential to the initial state (zero). Over time, these mechanisms are repeatedly carried out in each IF neuron, enabling event-based computations throughout the SNN layers.

\subsection{Spiking input event representation}
\label{sec:representation}
An event-based camera tracks the changes in log-scale intensity ($I$) at every element in the  pixel-array independently and generates a discrete event whenever the change exceeds a threshold ($\theta$):
\begin{equation}
\|\log(I_{t+1}) - \log(I_{t})\| \geq \theta 
\label{eq2}
\end{equation}
A discrete event contains a 4-tuple \{$x, y, t, p$\}, consisting of the coordinates: $x,y;$ timestamp: $t;$ and polarity (direction) of brightness change: $p$. This input representation is called Address Event Representation (AER), and is the standard format used by event-based sensors.

\begin{figure*}[ht]
\begin{center}
\includegraphics[width=0.66\textwidth]{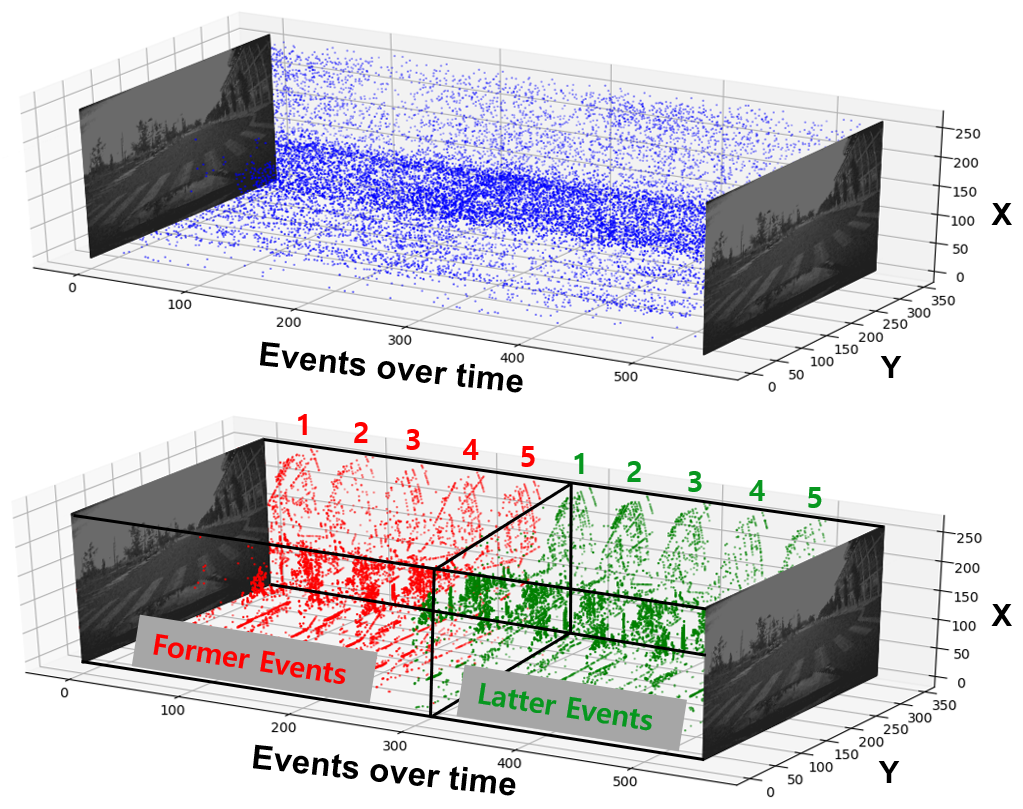}
\caption{Input event representation. ($Top$) Continuous raw events between two consecutive grayscale images from an event camera. ($Bottom$) Accumulated event frames between two consecutive grayscale images to form the former and the latter event groups, serving as inputs to the network.}
\label{fig2}
\vspace{-8mm}
\end{center}
\end{figure*}

There are prior works that have modified the representations of asynchronous event camera outputs to be compatible with ANN-based methods. To overcome the asynchronous nature, event outputs are typically recorded for a certain time period and transformed into a synchronous image-like representation. In  EV-FlowNet~\cite{zhu2018ev}, the most recent pixel-wise timestamps and the event counts encoded the motion information (within a time window) in an image. However, fast motions and dense events (in local regions of the image) can vastly overlap per-pixel timestamp information, and temporal information can be lost. In addition, \cite{zhu2019unsupervised} proposed a discretized event volume that deals with the time domain as a channel to retain the spatio-temporal event distributions. However, the number of input channels increases significantly as the time dimensions are finely discretized, further aggravating the computation and parameter overheads.

In this work, we propose a discretized input representation (fine-grained in time) that preserves the spatial and temporal information of events for SNNs. Our proposed input encoding scheme discretizes the time dimension within a time window into two groups (former and latter). Each group contains $N$ number of event frames obtained by accumulating raw events from the timestamp of the previous frame till the current timestamp. 
Each of these event frames is also composed of two channels for ON/OFF polarity of events. Hence, the input to the network consists of a sequence of $N$ frames with four channels (one frame each from the former and the latter groups having two channels each). The proposed input representation is displayed in Fig.~\ref{fig2} for one channel (assuming the number of event frames in each group equals to five). The main characteristic of our proposed input event representation (compared to ANN-based methods) are as follows:

\begin{itemize}
\item Our spatio-temporal input representations encode only the presence of events over time, allowing asynchronous and event-based computations in SNNs. In contrast, ANN-based input representation often requires the timestamp and the event count images in separate channels.
\item In Spike-FlowNet, each event frame from the former and the latter groups sequentially passes through the network, thereby preserving and utilizing the spatial and temporal information over time. On the contrary, ANN-based methods feed-forward all input information to the network at once.
\end{itemize}

\subsection{Self-Supervised Loss}
The DAVIS camera \cite{dvs240} is a commercially available event-camera, which simultaneously provides synchronous grayscale images and asynchronous event streams. The number of available event-based camera datasets with annotated labels suitable for optical flow estimation is quite small, as compared to frame-based camera datasets. Hence, a self-supervised learning method that uses proxy labels from the recorded grayscale images \cite{jason2016back,zhu2018ev} is employed for training our Spike-FlowNet.

The overall loss incorporates a photometric reconstruction loss ($\mathcal{L}_{\text{photo}}$) and a smoothness loss ($\mathcal{L}_{\text{smooth}}$) \cite{jason2016back}. To evaluate the photometric loss within each time window, the network is provided with the former and the latter event groups and a pair of grayscale images, taken at the start and the end of the event time window ($I_t, I_{t+dt}$). The predicted optical flow from the network is used to warp the second grayscale image to the first grayscale image. The photometric loss $(\mathcal{L}_{\text{photo}})$ aims to minimize the discrepancy between the first grayscale image and the inverse warped second grayscale image. This loss uses the photo-consistency assumption that a pixel in the first image remains similar in the second frame mapped by the predicted optical flow.
The photometric loss is computed as follows:
\begin{multline}
\mathcal{L}_{\text{photo}}(u,v;I_t, I_{t+dt}) =  \sum_{x,y} \rho(I_t(x,y) - I_{t+dt}(x+u(x,y),~y+v(x,y)))  
\label{eq3}
\end{multline}
where, $I_t, I_{t+dt}$ indicate the pixel intensity of the first and second grayscale images, $u, v$ are the flow estimates in the horizontal and vertical directions, $\rho$ is the Charbonnier loss $\rho(x) = (x^2 + \eta^2)^r$, which is a generic loss used for outlier rejection in optical flow estimation~\cite{sun2014}. For our work, $r=0.45$ and $\eta=$1e-3 show the optimum results for the computation of photometric loss.

Furthermore, a smoothness loss $(\mathcal{L}_{\text{smooth}})$ is applied for enhancing the spatial collinearity of neighboring optical flow. The smoothness loss minimizes the difference in optical flow between neighboring pixels and acts as a regularizer on the predicted flow. It is computed as follows:
\begin{multline}
\mathcal{L}_{\text{smooth}}(u,v) = \frac{1}{HD} \sum_{j}^{H} \sum_{i}^{D} (\|u_{i,j}-u_{i+1,j}\| + \|u_{i,j}-u_{i,j+1}\| \\+ \|v_{i,j}-v_{i+1,j}\| + \|v_{i,j}-v_{i,j+1}\|)
\label{eq4}
\end{multline} 
where $H$ is the height and $D$ is the width of the predicted flow output. The overall loss is computed as the weighted sum of the photometric and smoothness loss:
\begin{equation}
    \mathcal{L}_{\text{total}} = \mathcal{L}_{\text{photo}} + \lambda \mathcal{L}_{\text{smooth}}
\label{eq5}
\end{equation}
\noindent where $\lambda$ is the weight factor.

\begin{figure*}[ht]
\begin{center}
\includegraphics[width=0.95\textwidth]{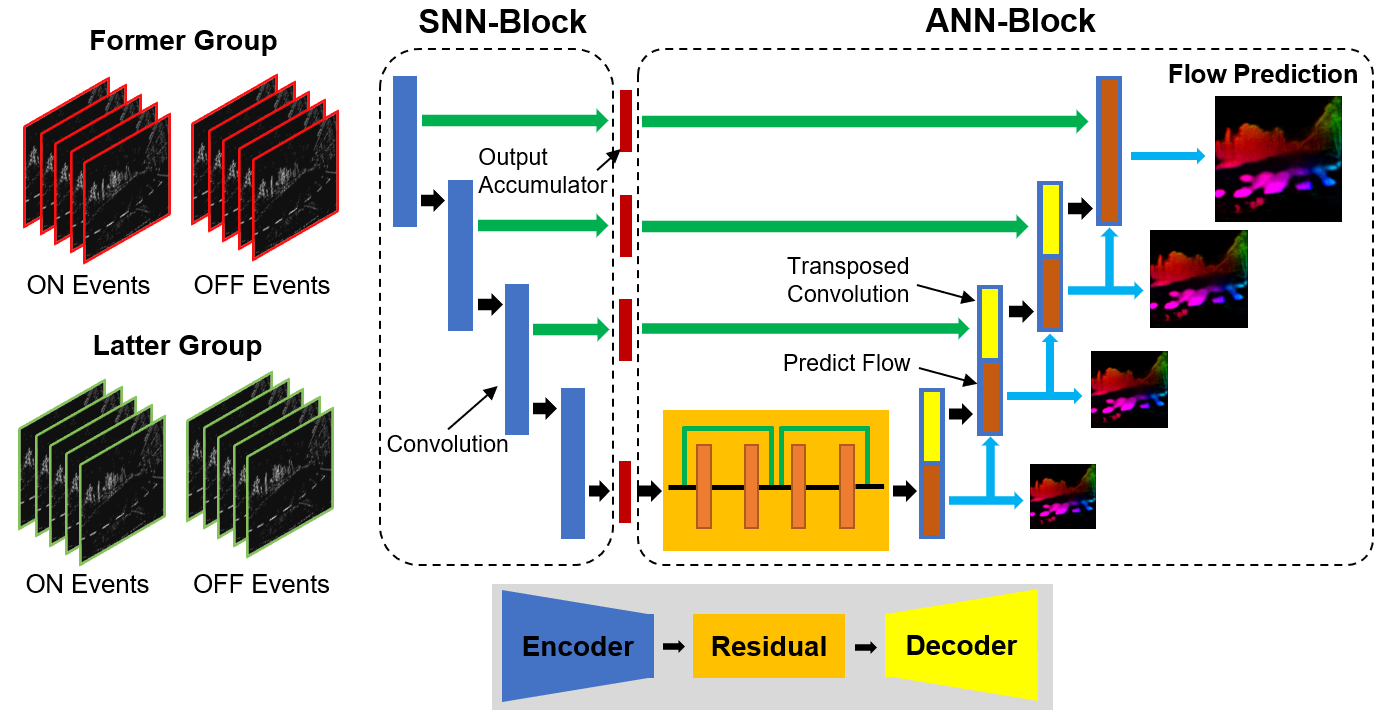}
\caption{Spike-FlowNet architecture. The four-channeled input images, comprised of ON/OFF polarity events for former and latter groups, are sequentially passed through the hybrid network. The SNN-block contains the encoder layers followed by output accumulators, while the ANN-block contains the residual and decoder layers. The loss is evaluated after forward propagating all consecutive input event frames (a total of $N$ inputs, sequentially taken in time from the former and the latter event groups) within the time window. The black arrows denote the forward path, green arrows represent residual connections, and blue arrows indicate the flow predictions.} 
\label{fig3}
\end{center}
\end{figure*}

\subsection{Spike-FlowNet Architecture}
Spike-FlowNet employs a deep hybrid architecture that accommodates SNNs and ANNs in different layers, enabling the benefits of SNNs for sparse event data processing and ANNs for maintaining the performance. The use of a hybrid architecture is attributed to the fact that spike activities reduce drastically with growing the network depth in the case of full-fledged SNNs. This is commonly referred to as the vanishing spike phenomenon \cite{panda2019towards}, and potentially leads to performance degradation in deep SNNs. Furthermore, high numerical precision is essentially required for estimating the accurate pixel-wise network outputs, namely the regression tasks. Hence, very rare and binary precision spike signals (in input and intermediate layers) pose a crucial issue for predicting the accurate flow displacements. 
To resolve these issues, only the encoder block is built as an SNN, while the residual and decoder blocks maintain an ANN architecture.

Spike-FlowNet's network topology resembles the U-Net~\cite{unet} architecture, containing four encoder layers, two residual blocks, and four decoder layers as shown in Fig.~\ref{fig3}. The events are represented as the four-channeled input frames as presented in Section~\ref{sec:representation}, and are sequentially passed through the SNN-based encoder layers over time (while being downsampled at each layer). Convolutions with a stride of two are employed for incorporating the functionality of dimensionality reduction in the encoder layers. The outputs from encoder layers are collected in their corresponding output accumulators until all consecutive event images have passed. Next, the accumulated outputs from final encoder layer are passed through two residual blocks and four decoder layers. The decoder layers upsample the activations using transposed convolution. At each decoder layer, there is a skip connection from the corresponding encoder layer, as well as another convolution layer to produce an intermediate flow prediction, which is concatenated with the activations from the transposed convolutions. The total loss is evaluated after the forward propagation of all consecutive input event frames through the network and is applied to each of the intermediate dense optical flows using the grayscale images.

\begin{algorithm}[ht]
   \caption{Backpropagation Training in Spike-FlowNet for an Iteration.}
   \label{alg:surrogate}
\begin{algorithmic}
   \STATE {\bfseries Input:} Event-based inputs ($inputs$), total number of discrete time-steps ($N$), number of SNN/ANN layers ($L_S/L_A$), SNN/ANN outputs ($o/o_A$) membrane potential ($V$), firing threshold ($V_{th}$), ANN nonlinearity ($f$)
   \STATE {\bfseries Initialize:} $V^l[n] = 0,~\forall l = 1,...,L_{S}$
   \STATE //~{\bfseries \textit{Forward Phase in SNN-blocks}}
   \FOR{$n \leftarrow 1$ {\bfseries to} $N$}
        \STATE $o^1[n] = inputs[n]$
        \FOR{$l \leftarrow 2$ {\bfseries to} $L_{S}-1$}         
            \STATE $V^l[n]~= V^l[n-1] + w^{l} o^{l-1}[n]$//\textit{weighted spike-inputs are integrated to $V$}
        \IF {$V^l[n] > V_{th}$}
            \STATE $o^{l}[n] = 1,~V^l[n] = 0$ //\textit{if $V$ exceeds $V_{th}$, a neuron emits a spike and reset V}
        \ENDIF
        \ENDFOR    
        \STATE $o_A^{L_S} = V^{L_S}[n]~= V^{L_S}[n-1] + w^{L_S} o^{L_S-1}[n]$ //\textit{final SNN layer does not fire}
   \ENDFOR
   \STATE //~{\bfseries \textit{Forward Phase in ANN-blocks}}
   \FOR{$l \leftarrow L_{S}+1$ {\bfseries to} $L_{S}+L_{A}$}         
        \STATE $o_A^{l}~= f(w^{l} o_A^{l-1})$
    \ENDFOR    
   \STATE //~{\bfseries \textit{Backward Phase in ANN-blocks}}
   \FOR{$l \leftarrow L_{S}+L_{A}$ {\bfseries to} $L_{S}$}    
   \STATE $\triangle w^l = \pderiv{\mathcal{L}_{total}}{o_A^{l}} \pderiv{o_A^{l}}{w^l}$
   \ENDFOR
   \STATE //~{\bfseries \textit{Backward Phase in SNN-blocks}}
   \FOR{$n \leftarrow N$ {\bfseries to} $1$}
       \FOR{$l \leftarrow L_S-1$ {\bfseries to} $1$}    
       \STATE //\textit{evaluate partial derivatives of loss w.r.t. $w_S$ by unrolling the SNN over time}
       \STATE $\triangle w^l[n] = \pderiv{\mathcal{L}_{total}}{o^{l}[n]} \pderiv{o^{l}[n]}{V^{l}[n]} \pderiv{V^{l}[n]}{w^l[n]}$
       \ENDFOR
   \ENDFOR
\end{algorithmic}
\end{algorithm}

\subsection{Backpropagation Training in Spike-FlowNet}

The spike generation function of an IF neuron is a hard threshold function that emits a spike when the membrane potential exceeds a firing threshold. Due to this discontinuous and non-differentiable neuron model, standard backpropagation algorithms cannot be applied to SNNs in their native form. Hence, several approximate methods have been proposed to estimate the surrogate gradient of spike generation function. In this work, we adopt the approximate gradient method proposed in \cite{lee2020enabling,lee2016training} for back-propagating errors through SNN layers. The approximate IF gradient is computed as $\frac{1}{V_{th}}$, where the threshold value accounts for the change of the spiking output with respect to the input. Algorithm \ref{alg:surrogate} illustrates the forward and backward pass in ANN-block and SNN-block.

In the forward phase, neurons in the SNN layers accumulate the weighted sum of the spike inputs in membrane potential. If the membrane potential exceeds a threshold, a neuron emits a spike at its output and resets. The final SNN layer neurons just integrate the weighted sum of spike inputs in the output accumulator, while not producing any spikes at the output. At the last time-step, the integrated outputs of SNN layers propagate to the ANN layers to predict the optical flow. After the forward pass, the final loss ($\mathcal{L}_{total}$) is evaluated, followed by backpropagation of gradients through the ANN layers using standard backpropagation. 

\begin{figure}[t]
\begin{center}
\includegraphics[width=0.7\columnwidth]{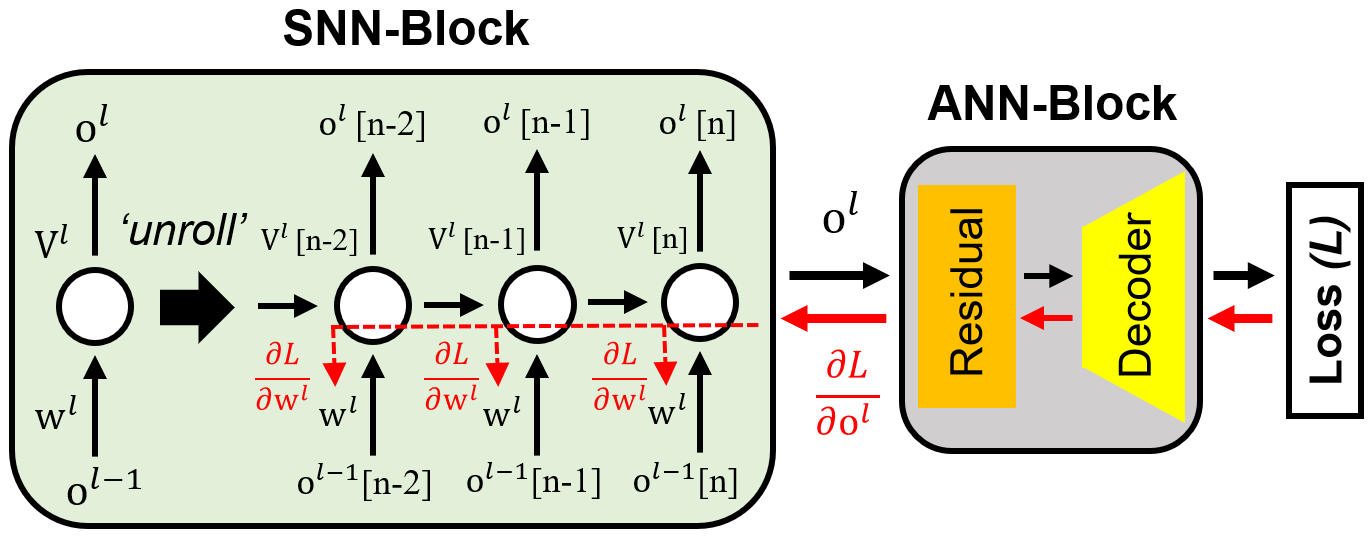}
\caption{Error backpropagation in Spike-FlowNet. After the forward pass, the gradients are back-propagated through the ANN block using standard backpropagation whereas the backpropagated errors ($\pderiv{\mathcal{L}}{o^l}$) pass through the SNN layers using the approximate IF gradient method and BPTT technique.} 
\label{fig4}
\end{center}
\end{figure}

Next, the backpropagated errors ($\pderiv{\mathcal{L}_{total}}{o^{L_S}}$) pass through the SNN layers using the approximate IF gradient method and BackPropagation Through Time (BPTT) \cite{werbos1990backpropagation}. In BPTT, the network is unrolled for all discrete time-steps, and the weight update is computed as the sum of gradients from each time-step. This procedure is displayed in Fig.~\ref{fig4} where the final loss is back-propagated through an ANN-block and a simple SNN-block consisting of a single input IF neuron. The parameter updates of the $l^{th}$ SNN layers are described as follows:
\begin{align}
    \triangle w^l = \sum_{n} \pderiv{\mathcal{L}_{total}}{o^l[n]} \pderiv{o^l[n]}{V^l[n]} \pderiv{V^l[n]}{w^l} \text{,~where~}\pderiv{o^l[n]}{V^l[n]}=\frac{1}{V_{th}}(o^l[n]>0)
\label{eq6}
\end{align}
where $o^l$ represents the output of spike generation function. This method enables the end-to-end self-supervised training in the proposed hybrid architecture.

\section{Experimental Results}
\subsection{Dataset and Training Details}
We use the MVSEC dataset~\cite{zhu2018multivehicle} for training and evaluating the optical flow predictions. MVSEC contains stereo event-based camera data for a variety of environments (e.g., indoor flying and outdoor driving) and also provides the corresponding ground truth optical flow. In particular, the indoor and outdoor sequences are recorded in dissimilar environments where the indoor sequences (indoor\_flying) have been captured in a lab environment and the outdoor sequences (outdoor\_day) have been recorded while driving on public roads.

Even though the indoor\_flying and outdoor\_day scenes are quite different, we only use outdoor\_day2 sequence for training Spike-FlowNet. This is done to provide fair comparisons with prior works~\cite{zhu2018ev,zhu2019unsupervised} which utilized only outdoor\_day2 sequence for training. During training, input images are randomly flipped horizontally and vertically (with 0.5 probability) and randomly cropped to $256 \times 256$ size. Adam optimizer \cite{kingma2014adam} is used, with the initial learning rate of 5e-5, and scaled by $0.7$ every $5$ epochs until $10$ epoch, and every $10$ epochs thereafter. The model is trained on the left event camera data of outdoor\_day2 sequence for $100$ epochs with a mini-batch size $8$. Training is done for two different time windows lengths (i.e, 1 grayscale image frame apart $(dt=1)$ and 4 grayscale image frames apart $(dt=4)$). The number of event frame ($N$) and weight factor for the smoothness loss ($\lambda$) are set to 5, 10 for a $dt=1$ case and 20, 1 for a $dt=4$ case, respectively. The threshold of the IF neurons are set to $0.5$ $(dt=4)$ and $0.75$ $(dt=1)$ in SNN layers.

\subsection{Algorithm Evaluation Metric}
The evaluation metric for optical flow prediction is the Average End-point Error (AEE), which represents the mean distance between the predicted flow ($y_{\text{pred}}$) and the ground truth flow ($y_{\text{gt}}$). It is given by:

\begin{equation}
\text{AEE} = \frac{1}{m} \sum_{\text{m}} \left\Vert(u,v)_{\text{pred}}-(u,v)_{\text{gt}}\right\Vert_2
\label{eq7}
\end{equation}

\noindent where $m$ is the number of active pixels in the input images. Because of the highly sparse nature of input events, the optical flows are only estimated at pixels where both the events and ground truth data is present. We compute the AEE for $dt=1$ and $dt=4$ cases.

\begin{table}[b]
\caption{Average Endpoint Error (AEE) comparisons with Zhu et al.~\cite{zhu2019unsupervised} and EV-FlowNet~\cite{zhu2018ev}.}
\begin{center}
\resizebox{\textwidth}{!}{
\begin{tabular}{llccccccccc}
\hline
   &  & \multicolumn{4}{c}{dt=1 frame} &  & \multicolumn{4}{c}{dt=4 frame} \\ \cline{3-6} \cline{8-11} 
  &  & indoor1  & indoor2 & indoor3 & outdoor1 &  & indoor1  & indoor2 & indoor3 & outdoor1 \\ \cline{1-1} \cline{3-6} \cline{8-11} 
Zhu et al. \cite{zhu2019unsupervised}              &  & \bf{0.58}     & \bf{1.02}    & \bf{0.87}    & \bf{0.32}     &  & \bf{2.18}     & 3.85    & \bf{3.18}    & 1.30     \\
EV-FlowNet \cite{zhu2018ev}               &  & 1.03     & 1.72    & 1.53    & 0.49     &  & 2.25     & 4.05    & 3.45    & 1.23     \\
This work                &  & 0.84     & 1.28    & 1.11    & 0.49     &  & 2.24     & \bf{3.83}    & \bf{3.18}    & \bf{1.09}     \\\hline

\end{tabular}}
* EV-FlowNet also uses a self-supervised learning method, providing the the fair comparison baseline compared to Spike-FlowNet.
\end{center}
\label{table1}
\end{table}

\begin{figure}[t]
\begin{center}
\includegraphics[width=1.02\textwidth]{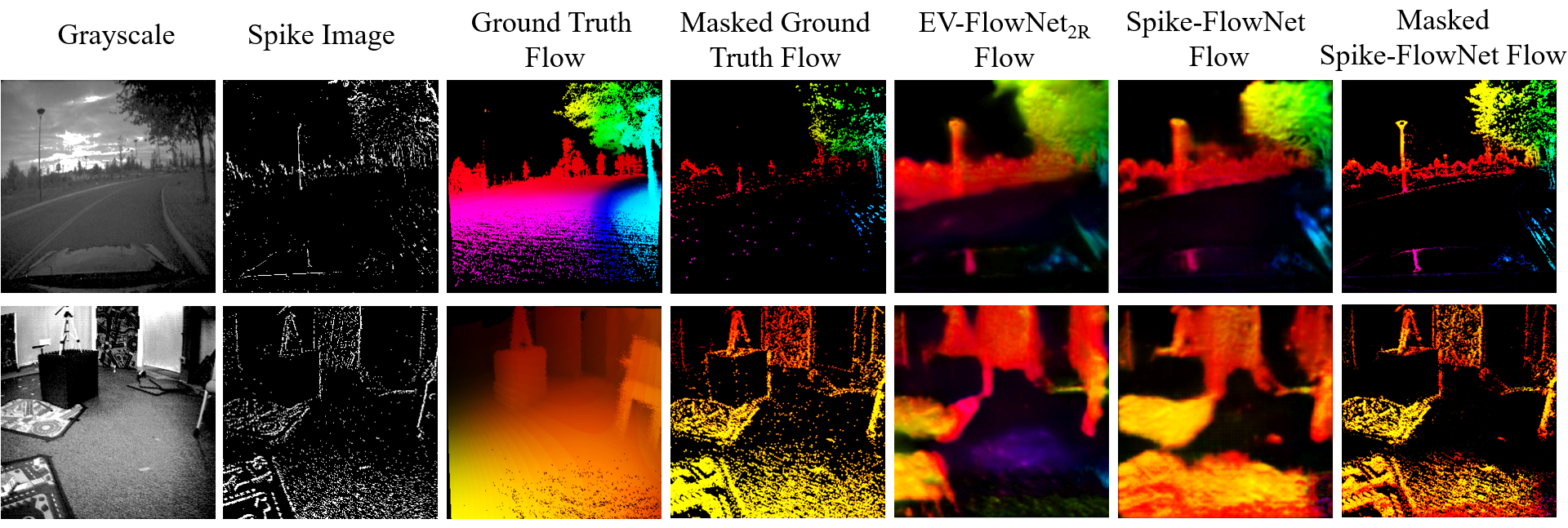}
\caption{Optical flow evaluation and comparison with EV-FlowNet. The samples are taken from ($top$) $outdoor\_day1$ and ($bottom$) $indoor\_day1$. The Masked Spike-FlowNet Flow is basically a sparse optical flow computed at pixels at which events occurred. It is computed by masking the predicted optical flow with the spike image.}
\label{fig5}
\end{center}
\end{figure}

\subsection{Average End-point Error (AEE) Results}
During testing, optical flow is estimated on the center cropped $256 \times 256$ left camera images of the indoor\_flying 1,2,3 and outdoor\_day 1 sequences. We use all events for the indoor\_flying sequences, but we take events within 800 grayscale frames for the outdoor\_day1 sequence, similar to~\cite{zhu2018ev}. Table \ref{table1} provides the AEE evaluation results in comparison with the prior event camera based optical flow estimation works. Overall, our results show that Spike-FlowNet can accurately predict the optical flow in both the indoor\_flying and outdoor\_day1 sequences. This demonstrates that the proposed Spike-FlowNet can generalize well to distinctly different environments. The grayscale, spike event, ground truth flow and the corresponding predicted flow images are visualized in Fig. \ref{fig5} where the images are taken from ($top$) $outdoor\_day1$ and ($bottom$) $indoor\_day1$, respectively. Since event cameras work based on changing light intensity at pixels, the regions having low texture produce very sparse events due to minimal intensity changes, resulting in scarce optical flow predictions in the corresponding areas such as the flat surfaces. Practically, the useful flows are extracted by using flow estimations at points where significant events exist in the input frames. 

Moreover, we compare our quantitative results with the recent works \cite{zhu2018ev,zhu2019unsupervised} on event-based optical flow estimation, as listed in Table \ref{table1}. We observe that Spike-FlowNet outperforms EV-FlowNet~\cite{zhu2018ev} in terms of AEE results in both the $dt=1$ and $dt=4$ cases. It is worth noting here that EV-FlowNet employs a similar network architecture and self-supervised learning method, providing a fair comparison baseline for fully ANN architectures. In addition, Spike-FlowNet attains AEE results slightly better or comparable to \cite{zhu2019unsupervised} in the $dt=4$ case, while underperforming in the $dt=1$ case. \cite{zhu2019unsupervised} presented an image deblurring based unsupervised learning that employed only the event streams. Hence, it seems to not suffer from the issues related to grayscale images such as motion blur or aperture problems during training. In view of these comparisons, Spike-FlowNet (with presented spatio-temporal event representation) is more suitable for motion detection when the input events have a certain minimum level of spike density. We further provide the ablation studies for exploring the optimal design choices in the supplementary material.

\begin{table}[b]
\begin{center}
\caption{Analysis for Spike-FlowNet in terms of the mean spike activity, the total and normalized number of SNN operations in an encoder-block, the encoder-block and overall computational energy benefits.}
\resizebox{\textwidth}{!}{
\begin{tabular}{ccccccccccccc}
\hline
                                 &  & \multicolumn{2}{c}{indoor1} &  & \multicolumn{2}{c}{indoor2} &  & \multicolumn{2}{c}{indoor3} &  & \multicolumn{2}{c}{outdoor1} \\ \cline{3-4} \cline{6-7} \cline{9-10} \cline{12-13} 
                                 &  & dt=1         & dt=4         &  & dt=1         & dt=4         &  & dt=1         & dt=4         &  & dt=1          & dt=4         \\ \cline{1-1} \cline{3-4} \cline{6-7} \cline{9-10} \cline{12-13} 
Encoder Spike Activity (\%)           &  & \textbf{0.33}         & 0.87         &  & 0.65         & 1.27         &  & 0.53         & 1.11         &  & 0.41          & 0.78         \\
Encoder SNN \# Operation ($\times10^8$)         &  & \textbf{0.16}         & 1.69         &  & 0.32         & 2.47         &  & 0.26         & 2.15         &  & 0.21          & 1.53         \\
Encoder Normalized \# Operation (\%)               &  & \textbf{1.68}         & 17.87        &  & 3.49         & 26.21        &  & 2.81         & 22.78        &  & 2.29          & 16.23        \\
Encoder Compute-energy Benefit ($\times$)   &  & \textbf{305}          & 28.6         &  & 146.5        & 19.5         &  & 182.1        & 22.44        &  & 223.2         & 31.5         \\
Overall Compute-energy Reduction (\%) &  & \textbf{17.57}        & 17.01        &  & 17.51        & 16.72        &  & 17.53        & 16.84        &  & 17.55         & 17.07        \\ \hline
\end{tabular}}
* For an ANN, the number of synaptic operations is $9.44\times10^8$ for the encoder-block and $5.35\times10^9$ for overall network.
\end{center}
\label{table2}
\end{table}

\subsection{Computational Efficiency}
To further analyze the benefits of Spike-FlowNet, we estimate the gain in computational costs compared to a fully ANN architecture. Typically, the number of synaptic operations is used as a metric for benchmarking the computational energy of neuromorphic hardware \cite{lee2020enabling,merolla2014million,rueckauer2017conversion}. Also, the required energy consumption per synaptic operation needs to be considered. Now, we describe the procedures for measuring the computational costs in SNN and ANN layers.

In a neuromorphic hardware, SNNs carry out event-based computations only at the arrival of input spikes. Hence, we first measure the mean spike activities at each time-step in the SNN layers. As presented in the first row of Table 2, the mean spiking activities (averaged over indoor1,2,3 and outdoor1 sequences) are $0.48\%$ and $1.01\%$ for $dt=1$ and $dt=4$ cases, respectively. Note that the neuronal threshold is set to a higher value in $dt=1$ case; hence the average spiking activity becomes sparser compared to $dt=4$ case. The extremely rare mean input spiking activities are mainly due to the fact that event camera outputs are highly sparse in nature. This sparse firing rate is essential for exploiting efficient event-based computations in SNN layers. In contrast, ANNs execute dense matrix-vector multiplication operations without considering the sparsity of inputs. In other words, ANNs simply feed-forward the inputs at once, and the total number of operations are fixed. This leads to the high energy requirements (compared to SNNs) by computing both zero and non-zero entities, especially when inputs are very sparse.

Essentially, SNNs need to compute the spatio-temporal spike images over a number of time-steps. Given $M$ is the number of neurons, $C$ is number of synaptic connections and $F$ indicates the mean firing activity, the number of synaptic operations at each time-step in the $l^{th}$ layer is calculated as $M_l \times C_l \times F_l$. The total number of SNN operations is the summation of synaptic operations in SNN layers during the $N$ time-steps. Hence, the total number of SNN and ANN operations become $\sum_{l}(M_l \times C_l \times F_l) \times N$ and $\sum_{l} M_l \times C_l$, respectively. Based on these, we estimate and compare the average number of synaptic operations on Spike-FlowNet and a fully ANN architecture. The total and the normalized number of SNN operations compared to ANN operations on the encoder-block are provided in the second and the third row of Table 2, respectively. 

Due to the binary nature of spike events, SNNs perform only accumulation (AC) per synaptic operation. On the other hand, ANNs perform the multiply-accumulate (MAC) computations since the inputs consist of analog-valued entities. In general, AC computation is considered to be significantly more energy-efficient than MAC. For example, AC is reported to be $5.1\times$ more energy-efficient than a MAC in the case of 32-bit floating-point numbers (45nm CMOS process)~\cite{horowitz20141}. Based on this principle, the computational energy benefits of encoder-block and overall Spike-FlowNet are obtained, as provided in the fourth and the fifth rows of Table 2, respectively. These results reveal that the SNN-based encoder-block is 214.2$\times$ and 25.51$\times$ more computationally efficient compared to ANN-based one (averaged over indoor1,2,3 and outdoor1 sequences) for $dt=1$ and $dt=4$ cases, respectively. The number of time-steps ($N$) is four times less in $dt=1$ case than in $dt=4$ case; hence, the computational energy benefit is much higher in $dt=1$ case. 

From our analysis, the proportion of required computations in encoder-block compared to the overall architecture is $17.6\%$. This reduces the overall energy benefits of Spike-FlowNet. In such a case, an approach of interest would be to perform a distributed edge-cloud implementation where the SNN- and ANN-blocks are administered on the edge device and the cloud, respectively. This would lead to high energy benefits on edge devices, which are limited by resource constraints while not compromising on algorithmic performance.

\section{Conclusion}
In this work, we propose Spike-FlowNet, a deep hybrid architecture for energy-efficient optical flow estimations using event camera data. To leverage the benefits of both SNNs and ANNs, we integrate them in different layers for resolving the spike vanishing issue in deep SNNs. Moreover, we present a novel input encoding strategy for handling outputs from event cameras, preserving the spatial and temporal information over time. Spike-FlowNet is trained with a self-supervised learning method, bypassing expensive labeling. The experimental results show that the proposed architecture accurately predicts the optical flow from discrete and asynchronous event streams along with substantial benefits in terms of computational efficiency compared to the corresponding ANN architecture. 

\section*{Acknowledgment}
This work was supported in part by C-BRIC, one of six centers in JUMP, a Semiconductor Research Corporation (SRC) program sponsored by DARPA, the National Science Foundation, Sandia National Laboratory, and the DoD Vannevar Bush Fellowship.

\section{Ablation Study}
In this supplementary material, we present the ablation studies to explore the optimal design choices of hybrid networks, input data representation and weight factor ($\lambda$) of the smoothness loss in the loss function.

\subsection{Hybrid Network}

In addition to the described architecture (denoted  Spike-FlowNet), we train additional network topologies to test different hybrid design options. We use two more networks in which residual blocks are composed of SNN layers: one where only first residual block is converted to SNN (Spike-FlowNet\_1R), and second where both residual blocks are converted to SNN (Spike-FlowNet\_2R). Note, results for a fully ANN architecture are given in EV-FlowNet \cite{zhu2018ev}. We do not consider converting the decoder layers to construct a fully SNN architecture, as they use analog inputs from intermediate optical flows and output accumulators.

Rows 1-3 in table \ref{table3} show the AEE results for the different network topologies. 
We find that AEE results degrade as more layers are transferred to SNNs for both $dt=1$ and $dt=4$. This is because the spike vanishing phenomenon aggravates with the network depth, leading to the degradation in the quality of predicted optical flow. The best AEE results are achieved by Spike-FlowNet case which is advocated throughout the manuscript. 

\subsection{Input representation}

We validate the influence of the number of groups ($N$) in input representation. In the case of $N=3$ and $N=4$, AEE results are provided in rows 4-5 in table \ref{table3}. Note, Spike-FlowNet represents $N=2$ case. With the increase in the number of input groups ($N$), the results show that $dt=1$ case achieves worse AEE while $dt=4$ converges to a reasonably accurate flow estimate. This is because each input group requires to have a certain number of events for proper training, and we find that $N=2$ case provides optimal results for both $dt=1$ and $dt=4$.

\begin{table}[h]
\caption{Average Endpoint Error (AEE) for ablation studies with different design choices}
\vspace{-6mm}
\begin{center}
\resizebox{\textwidth}{!}{
\begin{tabular}{ccccccccccc}
\hline
   &  & \multicolumn{4}{c}{dt=1 frame} &  & \multicolumn{4}{c}{dt=4 frame} \\ \cline{3-6} \cline{8-11} 
  &  & indoor1  & indoor2 & indoor3 & outdoor1 &  & indoor1  & indoor2 & indoor3 & outdoor1 \\ \cline{1-1} \cline{3-6} \cline{8-11} 
Spike-FlowNet                   &  & 0.84     & 1.28    & 1.11    & 0.49     &  & 2.24     & 3.83    & 3.18    & 1.09     \\
Spike-FlowNet\_1R                &  & 0.88     & 1.55    & 1.31    & 0.51     &  & 2.73     & 4.46    & 3.66    & 1.15     \\
Spike-FlowNet\_2R                &  & 0.90     & 1.56    & 1.29    & 0.56     &  & 2.75     & 4.61    & 3.76    & 1.19     \\ \cline{1-1} \cline{3-6} \cline{8-11} 
$N$=3                &  & 0.92     & 1.34    & 1.18    & 0.50     &  & 2.34     & 4.05    & 3.29    & 1.12     \\
$N$=4                &  & 1.07     & 1.76    & 1.57    & 0.60     &  & 2.27     & 3.81    & 3.10    & 1.15     \\ \cline{1-1} \cline{3-6} \cline{8-11} 
$\lambda$=1                &  & 0.91     & 1.38    & 1.23    & 0.50     &  & 2.24     & 3.83    & 3.18    & 1.09     \\
$\lambda$=10                &  & 0.84     & 1.28    & 1.11    & 0.49     &  & 2.42     & 4.22    & 3.44    & 1.18     \\
$\lambda$=100                &  & 0.84     & 1.30    & 1.14    & 0.49     &  & 2.50     & 4.01    & 3.28    & 1.19     \\\hline
\end{tabular}}
\end{center}
\label{table3}
\vspace{-6mm}
\end{table}

\subsection{Loss function}

To find the optimal ratio between photometric and smoothness losses, we train networks with a variety of weight factors ($\lambda$) over the range $[1, 100]$. Rows 6-8 in table \ref{table3} highlight AEE results for $\lambda$ = 1, 10, 100. We observe that $\lambda$ = 10, 100 cases converge to more accurate flow estimate for $dt=1$ while $\lambda$ = 1 case works better for $dt=4$. This is because inputs are greatly sparse in $dt=1$, hence its corresponding flow outputs have more scarce and discontinuous structures, requiring a higher degree of smoothness.


%
%
\bibliographystyle{splncs04}
\bibliography{egbib}

\begin{thebibliography}{10}
\providecommand{\url}[1]{\texttt{#1}}
\providecommand{\urlprefix}{URL }
\providecommand{\doi}[1]{https://doi.org/#1}

\bibitem{aung2018}
{Aung}, M.T., {Teo}, R., {Orchard}, G.: Event-based plane-fitting optical flow
  for dynamic vision sensors in fpga. In: 2018 IEEE International Symposium on
  Circuits and Systems (ISCAS). pp.~1--5 (May 2018).
  \doi{10.1109/ISCAS.2018.8351588}

\bibitem{barranco2015}
Barranco, F., Fermuller, C., Aloimonos, Y.: Bio-inspired motion estimation with
  event-driven sensors. In: Rojas, I., Joya, G., Catala, A. (eds.) Advances in
  Computational Intelligence. pp. 309--321. Springer International Publishing,
  Cham (2015)

\bibitem{benosman2}
{Benosman}, R., {Clercq}, C., {Lagorce}, X., {Ieng}, S., {Bartolozzi}, C.:
  Event-based visual flow. IEEE Transactions on Neural Networks and Learning
  Systems  \textbf{25}(2),  407--417 (Feb 2014).
  \doi{10.1109/TNNLS.2013.2273537}

\bibitem{benosman1}
Benosman, R., Ieng, S.H., Clercq, C., Bartolozzi, C., Srinivasan, M.:
  Asynchronous frameless event-based optical flow. Neural Networks
  \textbf{27},  32 -- 37 (2012).
  \doi{https://doi.org/10.1016/j.neunet.2011.11.001},
  \url{http://www.sciencedirect.com/science/article/pii/S0893608011002930}

\bibitem{flymotion2010}
Borst, A., Haag, J., Reiff, D.F.: Fly motion vision. Annual Review of
  Neuroscience  \textbf{33}(1),  49--70 (2010).
  \doi{10.1146/annurev-neuro-060909-153155},
  \url{https://doi.org/10.1146/annurev-neuro-060909-153155}, pMID: 20225934

\bibitem{dvs240}
{Brandli}, C., {Berner}, R., {Yang}, M., {Liu}, S., {Delbruck}, T.: A 240 ×
  180 130 db 3 µs latency global shutter spatiotemporal vision sensor. IEEE
  Journal of Solid-State Circuits  \textbf{49}(10),  2333--2341 (Oct 2014).
  \doi{10.1109/JSSC.2014.2342715}

\bibitem{brosch2015}
Brosch, T., Tschechne, S., Neumann, H.: On event-based optical flow detection.
  Frontiers in neuroscience  \textbf{9}, ~137 (04 2015).
  \doi{10.3389/fnins.2015.00137}

\bibitem{burkitt2006review}
Burkitt, A.N.: A review of the integrate-and-fire neuron model: I. homogeneous
  synaptic input. Biological cybernetics  \textbf{95}(1),  1--19 (2006)

\bibitem{loihi2018}
{Davies}, M., {Srinivasa}, N., {Lin}, T., {Chinya}, G., {Cao}, Y., {Choday},
  S.H., {Dimou}, G., {Joshi}, P., {Imam}, N., {Jain}, S., {Liao}, Y., {Lin},
  C., {Lines}, A., {Liu}, R., {Mathaikutty}, D., {McCoy}, S., {Paul}, A.,
  {Tse}, J., {Venkataramanan}, G., {Weng}, Y., {Wild}, A., {Yang}, Y., {Wang},
  H.: Loihi: A neuromorphic manycore processor with on-chip learning. IEEE
  Micro  \textbf{38}(1),  82--99 (January 2018).
  \doi{10.1109/MM.2018.112130359}

\bibitem{dayan2001theoretical}
Dayan, P., Abbott, L.F.: Theoretical neuroscience, vol.~806. Cambridge, MA: MIT
  Press (2001)

\bibitem{diehl2015unsupervised}
Diehl, P.U., Cook, M.: Unsupervised learning of digit recognition using
  spike-timing-dependent plasticity. Frontiers in computational neuroscience
  \textbf{9}, ~99 (2015)

\bibitem{gallego2018}
Gallego, G., Rebecq, H., Scaramuzza, D.: A unifying contrast maximization
  framework for event cameras, with applications to motion, depth, and optical
  flow estimation. CoRR  \textbf{abs/1804.01306} (2018),
  \url{http://arxiv.org/abs/1804.01306}

\bibitem{haessig2018spiking}
Haessig, G., Cassidy, A., Alvarez, R., Benosman, R., Orchard, G.: Spiking
  optical flow for event-based sensors using ibm's truenorth neurosynaptic
  system. IEEE transactions on biomedical circuits and systems  \textbf{12}(4),
   860--870 (2018)

\bibitem{horowitz20141}
Horowitz, M.: 1.1 computing's energy problem (and what we can do about it). In:
  2014 IEEE International Solid-State Circuits Conference Digest of Technical
  Papers (ISSCC). pp. 10--14. IEEE (2014)

\bibitem{jason2016back}
Jason, J.Y., Harley, A.W., Derpanis, K.G.: Back to basics: Unsupervised
  learning of optical flow via brightness constancy and motion smoothness. In:
  European Conference on Computer Vision. pp. 3--10. Springer (2016)

\bibitem{kingma2014adam}
Kingma, D.P., Ba, J.: Adam: A method for stochastic optimization. arXiv
  preprint arXiv:1412.6980  (2014)

\bibitem{lai2017}
Lai, W.S., Huang, J.B., Yang, M.H.: Semi-supervised learning for optical flow
  with generative adversarial networks. In: Guyon, I., Luxburg, U.V., Bengio,
  S., Wallach, H., Fergus, R., Vishwanathan, S., Garnett, R. (eds.) Advances in
  Neural Information Processing Systems 30, pp. 354--364. Curran Associates,
  Inc. (2017),
  \url{http://papers.nips.cc/paper/6639-semi-supervised-learning-for-optical-flow-with-generative-adversarial-networks.pdf}

\bibitem{lee2020enabling}
Lee, C., Sarwar, S.S., Panda, P., Srinivasan, G., Roy, K.: Enabling spike-based
  backpropagation for training deep neural network architectures. Frontiers in
  Neuroscience  \textbf{14}, ~119 (2020)

\bibitem{lee2016training}
Lee, J.H., Delbruck, T., Pfeiffer, M.: Training deep spiking neural networks
  using backpropagation. Frontiers in neuroscience  \textbf{10}, ~508 (2016)

\bibitem{dvs128}
{Lichtsteiner}, P., {Posch}, C., {Delbruck}, T.: A 128$\times$ 128 120 db 15
  $\mu$s latency asynchronous temporal contrast vision sensor. IEEE Journal of
  Solid-State Circuits  \textbf{43}(2),  566--576 (Feb 2008).
  \doi{10.1109/JSSC.2007.914337}

\bibitem{liu2018}
Liu, M., Delbr{\"{u}}ck, T.: {ABMOF:} {A} novel optical flow algorithm for
  dynamic vision sensors. CoRR  \textbf{abs/1805.03988} (2018),
  \url{http://arxiv.org/abs/1805.03988}

\bibitem{lucaskanade}
Lucas, B.D., Kanade, T.: An iterative image registration technique with an
  application to stereo vision. In: Proceedings of the 7th International Joint
  Conference on Artificial Intelligence - Volume 2. pp. 674--679. IJCAI'81,
  Morgan Kaufmann Publishers Inc., San Francisco, CA, USA (1981),
  \url{http://dl.acm.org/citation.cfm?id=1623264.1623280}

\bibitem{meister2018unflow}
Meister, S., Hur, J., Roth, S.: Unflow: Unsupervised learning of optical flow
  with a bidirectional census loss. In: Thirty-Second AAAI Conference on
  Artificial Intelligence (2018)

\bibitem{merolla2014million}
Merolla, P.A., Arthur, J.V., Alvarez-Icaza, R., Cassidy, A.S., Sawada, J.,
  Akopyan, F., Jackson, B.L., Imam, N., Guo, C., Nakamura, Y., et~al.: A
  million spiking-neuron integrated circuit with a scalable communication
  network and interface. Science  \textbf{345}(6197),  668--673 (2014)

\bibitem{Orchard2013}
Orchard, G., Benosman, R.B., Etienne-Cummings, R., Thakor, N.V.: A spiking
  neural network architecture for visual motion estimation. 2013 IEEE
  Biomedical Circuits and Systems Conference (BioCAS) pp. 298--301 (2013)

\bibitem{panda2019towards}
Panda, P., Aketi, S.A., Roy, K.: Toward scalable, efficient, and accurate deep
  spiking neural networks with backward residual connections, stochastic
  softmax, and hybridization. Frontiers in Neuroscience  \textbf{14}, ~653
  (2020)

\bibitem{paredes2019unsupervised}
Paredes-Vall{\'e}s, F., Scheper, K.Y.W., De~Croon, G.C.H.E.: Unsupervised
  learning of a hierarchical spiking neural network for optical flow
  estimation: From events to global motion perception. IEEE transactions on
  pattern analysis and machine intelligence  (2019)

\bibitem{unet}
Ronneberger, O., Fischer, P., Brox, T.: U-net: Convolutional networks for
  biomedical image segmentation. CoRR  \textbf{abs/1505.04597} (2015),
  \url{http://arxiv.org/abs/1505.04597}

\bibitem{rueckauer2017conversion}
Rueckauer, B., Lungu, I.A., Hu, Y., Pfeiffer, M., Liu, S.C.: Conversion of
  continuous-valued deep networks to efficient event-driven networks for image
  classification. Frontiers in neuroscience  \textbf{11}, ~682 (2017)

\bibitem{sun2014}
Sun, D., Roth, S., Black, M.J.: A quantitative analysis of current practices in
  optical flow estimation and the principles behind them. Int. J. Comput.
  Vision  \textbf{106}(2),  115--137 (Jan 2014).
  \doi{10.1007/s11263-013-0644-x},
  \url{http://dx.doi.org/10.1007/s11263-013-0644-x}

\bibitem{werbos1990backpropagation}
Werbos, P.J.: Backpropagation through time: what it does and how to do it.
  Proceedings of the IEEE  \textbf{78}(10),  1550--1560 (1990)

\bibitem{zhu2017}
{Zhu}, A.Z., {Atanasov}, N., {Daniilidis}, K.: Event-based feature tracking
  with probabilistic data association. In: 2017 IEEE International Conference
  on Robotics and Automation (ICRA). pp. 4465--4470 (May 2017).
  \doi{10.1109/ICRA.2017.7989517}

\bibitem{zhu2018multivehicle}
Zhu, A.Z., Thakur, D., {\"O}zaslan, T., Pfrommer, B., Kumar, V., Daniilidis,
  K.: The multivehicle stereo event camera dataset: An event camera dataset for
  3d perception. IEEE Robotics and Automation Letters  \textbf{3}(3),
  2032--2039 (2018)

\bibitem{zhu2018ev}
Zhu, A.Z., Yuan, L., Chaney, K., Daniilidis, K.: Ev-flownet: Self-supervised
  optical flow estimation for event-based cameras. arXiv preprint
  arXiv:1802.06898  (2018)

\bibitem{zhu2019unsupervised}
Zhu, A.Z., Yuan, L., Chaney, K., Daniilidis, K.: Unsupervised event-based
  learning of optical flow, depth, and egomotion. In: Proceedings of the IEEE
  Conference on Computer Vision and Pattern Recognition. pp. 989--997 (2019)

\end{thebibliography}
\end{document}